\documentclass[10pt,twocolumn,letterpaper]{article}

\usepackage{iccv}

\usepackage{bbm}
\usepackage{multirow}
\usepackage[dvipsnames]{xcolor}
\usepackage{graphicx}
\usepackage{colortbl}
\usepackage{color}
\usepackage{amsmath,amssymb} 
\usepackage[american]{babel}
\usepackage{csquotes}
\usepackage{makecell}
\usepackage{pifont}

\definecolor{iccvblue}{rgb}{0.21,0.49,0.74}
\usepackage[pagebackref,breaklinks,colorlinks,allcolors=iccvblue]{hyperref}

\def\paperID{2904}
\def\confName{ICCV}
\def\confYear{2025}

\title{Learning Yourself: Class-Incremental Semantic Segmentation with Language-Inspired Bootstrapped Disentanglement}

\author{
  Ruitao Wu\textsuperscript{1,2} \quad
  Yifan Zhao\textsuperscript{1}\thanks{Corresponding authors.} \quad
  Jia Li\textsuperscript{1}\footnotemark[1]
  \\
  [0.4em] 
  \textsuperscript{1}State Key Laboratory of Virtual Reality Technology and Systems, SCSE \& QRI, Beihang University \\
  \textsuperscript{2}Zhongguancun Academy, Beijing, China \\
  {\tt\small \{ruitaowu, zhaoyf, jiali\}@buaa.edu.cn}
}

\begin{document}
\maketitle
\begin{abstract}

Class-Incremental Semantic Segmentation (CISS) requires continuous learning of newly introduced classes while retaining knowledge of past classes. By abstracting mainstream methods into two stages (visual feature extraction and prototype-feature matching), we identify a more fundamental challenge termed \textbf{catastrophic semantic entanglement}. This phenomenon involves Prototype-Feature Entanglement caused by semantic misalignment during the incremental process, and Background-Increment Entanglement due to dynamic data evolution. Existing techniques, which rely on visual feature learning without sufficient cues to distinguish targets, introduce significant noise and errors. To address these issues, we introduce a \textbf{L}anguage-inspired \textbf{B}ootstrapped \textbf{D}isentanglement framework (\textbf{LBD}). We leverage the prior class semantics of pre-trained visual-language models (e.g., CLIP) to guide the model in autonomously disentangling features through Language-guided Prototypical Disentanglement and Manifold Mutual Background Disentanglement. The former guides the disentangling of new prototypes by treating hand-crafted text features as topological templates, while the latter employs multiple learnable prototypes and mask-pooling-based supervision for background-incremental class disentanglement.
By incorporating soft prompt tuning and encoder adaptation modifications, we further bridge the capability gap of CLIP between dense and sparse tasks, achieving state-of-the-art performance on both Pascal VOC and ADE20k, particularly in multi-step scenarios.

\end{abstract}    
\section{Introduction}
\label{sec:intro}

Semantic segmentation~\cite{Deeplab,Segmenter,Mask2Former,Zhao2022FromPT} is a crucial computer vision task that assigns meaningful labels to each pixel in an image. While conventional networks perform well on datasets with all class labels available upfront, real-world applications require models that can adapt and incrementally learn new classes after deployment. Class-Incremental Semantic Segmentation (CISS)~\cite{ILT,SurveyCSS} is dedicated to solving this problem by enabling models to incorporate new classes through supervision while retaining the knowledge of previously learned classes.

\begin{figure}[t]
    \centering
    \includegraphics[width=\linewidth]{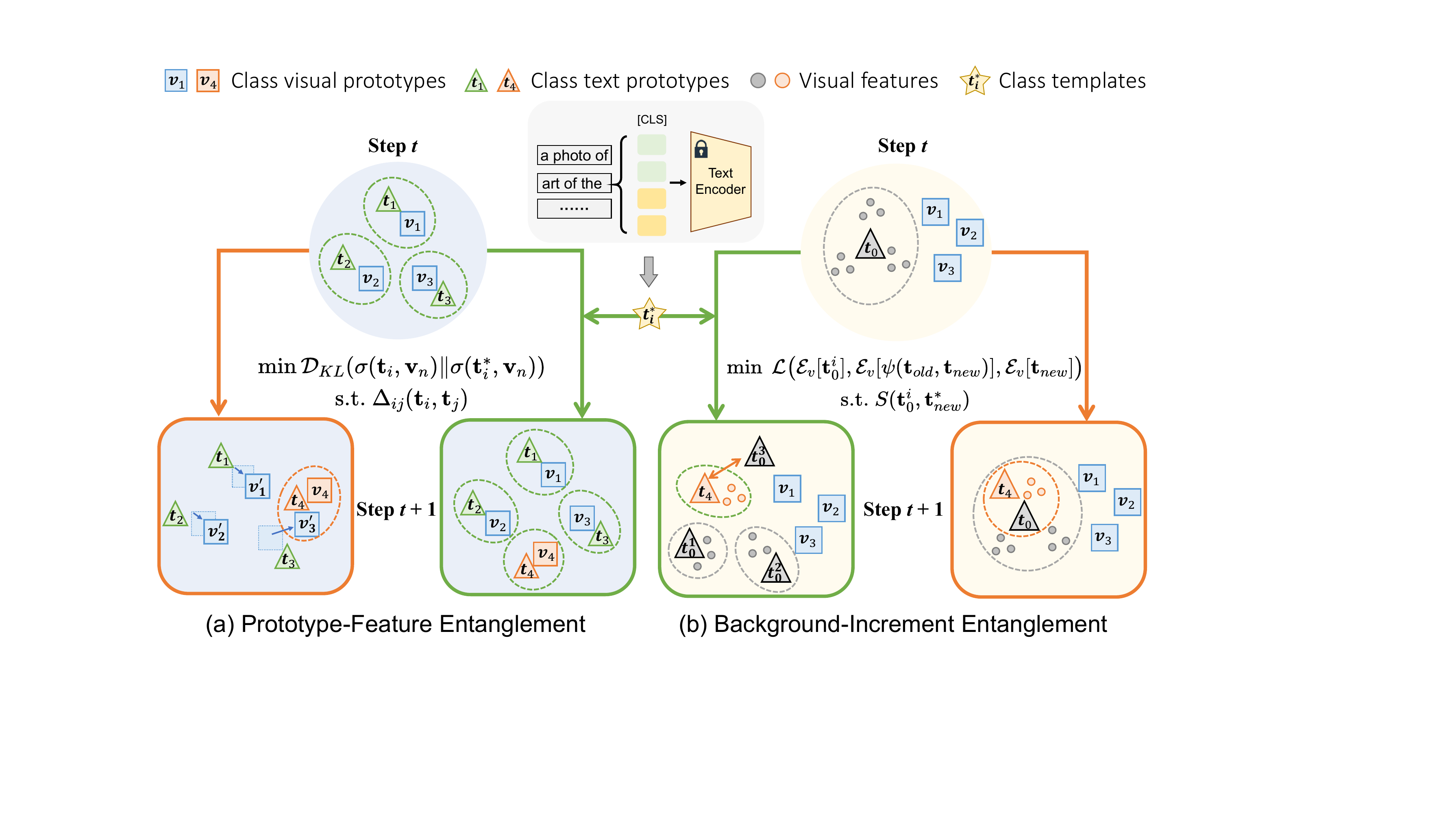}
    \caption{\textbf{Illustration of Catastrophic Semantic Entanglement (\textcolor{orange}{Orange}) and our countermeasures (\textcolor{teal}{Green}).} (a) Prototype-Feature Entanglement caused by the inter-class topology disruption. (b) Background-Increment Entanglement caused by the dynamically evolving foreground. We address the two issues through language-guided prototypical disentanglement (\cref{sec:Topology-Preserving Distillation Loss}) and manifold mutual background disentanglement (\cref{sec:Dynamic Multiple Background Representation}).}
    \label{fig:fig0}
    \vspace{-1em}
\end{figure}

A considerable amount of effort has been devoted to addressing the core issue of catastrophic forgetting in CISS from different perspectives. Data replay-based methods~\cite{RECALL,SDR,10.1145/3560071.3560080,tikp} store or generate old instances or features to review prior knowledge. Dynamic architecture-based methods~\cite{DKD,DER,Liu2020DynamicEN,Ye2022LearningWR,Yang2022DeepMR} benefit from flexible and scalable model frameworks. Recently, knowledge distillation-based strategies~\cite{Michieli2019IncrementalLT,Qiu2022SATSST,Phan2022ClassSW,Rong2023MiCroMC} have gained more attention due to their efficiency and simplicity, yet stabilizing the representation of new knowledge while preventing forgetting remains challenging.

To investigate the essence of this crux, we abstract mainstream methods into two key processes: visual feature extraction (\eg, pixel-level, mask-level) and matching predefined class prototypes (\eg, linear layers, query prompts). As new classes are added, this pipeline inevitably faces what we term \textit{catastrophic semantic entanglement}, which manifests in two aspects:
(i) \textbf{Prototype-Feature Entanglement} (\textcolor{orange}{Orange} box in \cref{fig:fig0}(a)). The inaccessibility of old data limits prototype differentiation to sparse training data, unable to address distribution overlap between similar classes and relationships between prototypes. This leads to semantic misalignment during the incremental process, exemplified by prototype confusion and feature overlap (\eg, prototype $\mathbf{t}_4$ entangling with $\mathbf{v}_3'$). Further training of the image encoder causes detrimental shifts in visual features. Since the model has not been fully learned, subsequent knowledge distillation exacerbates cumulative errors.
(ii) \textbf{Background-Increment Entanglement} (\textcolor{orange}{Orange} box in \cref{fig:fig0}(b)). The expansion of foreground classes continuously alters background semantics, removing new classes and incorporating background information from new datasets, which misaligns the background prototype with other classes (\eg, prototype $\mathbf{t}_4$ entangling with the background $\mathbf{t}_0$).

The commonality of these two issues lies in the reliance on visual representations alone to decouple entangled features. Although the visual module can self-discover beneficial feature distributions during training, it still lacks the necessary supervisory cues, leading to substantial noise and errors. Based on these observations, we propose the language-inspired bootstrapped disentanglement framework, aiming to guide the model in learning to bootstrap the decoupling of entangled features through prior class semantics from pre-trained visual-language models.

Specifically, for Prototype-Feature Entanglement, we design language-guided prototypical disentanglement (\textcolor{teal}{Green} box in \cref{fig:fig0}(a)). We treat manually constructed prompts with explicit class names as \textit{templates} containing generalized knowledge. By simultaneously minimizing the KL divergence between the patch-prototype and patch-template matching logits, as well as the KL divergence between the prototype-template distributions (represented by $\sigma$), we ensure topological stability at the macroscopic level. At the microscopic level, local plasticity is achieved through an orthogonal constraint (represented by $\Delta$) based on sorted scores.

For background entanglement, we design manifold mutual background disentanglement (\textcolor{teal}{Green} box in \cref{fig:fig0}(b)). The key to resolving background entanglement lies in \textit{eliminating} the semantic information of the current class embedded in the background from the previous step. We use CLIP score maps and ground truth to disentangle background reference features (operation represented by $\psi$) and apply supervised contrastive learning (loss function denoted by $\mathcal{L}$) to disentangle new class features from the background. Since background classes are composites of multiple semantics, we employ multiple learnable prototypes to represent the background, selecting the maximum activation value during class mask computation. Building on existing background weight transfer methods, we replicate weights based on the similarity between background embeddings and new class templates.

These two aspects enhance the model's continuous learning from different perspectives. However, the substantial modality gap in the original CLIP embedding space creates a large distance between image and text embeddings, limiting segmentation performance. Drawing inspiration from ~\cite{DenseCLIP}, we create CoOp-style~\cite{CoOP} text prompts for each class to derive corresponding text features. Additionally, we incorporate improvements from existing methods to further optimize CLIP's performance in dense scenarios.

In summary, the contributions of this paper are as follows:
\begin{itemize}
\item We propose an efficient framework for integrating CLIP into the CISS task with language-inspired bootstrapped disentanglement, which outperforms existing state-of-the-art methods on the Pascal VOC and ADE20k datasets.
\item To tackle the entanglement between class prototypes and visual features, we introduce language-guided prototypical disentanglement, which treats the vanilla text features as topological templates to guide the disentanglement of new prototypes.
\item To tackle the entanglement between background semantics and incremental classes, we introduce manifold mutual background disentanglement, which achieves the mutual disentanglement of the background and new classes through multiple learnable prompts and mask-pooling-based contrastive supervision.
\end{itemize}

\begin{figure*}[t]
    \centering
    \includegraphics[width=\linewidth]{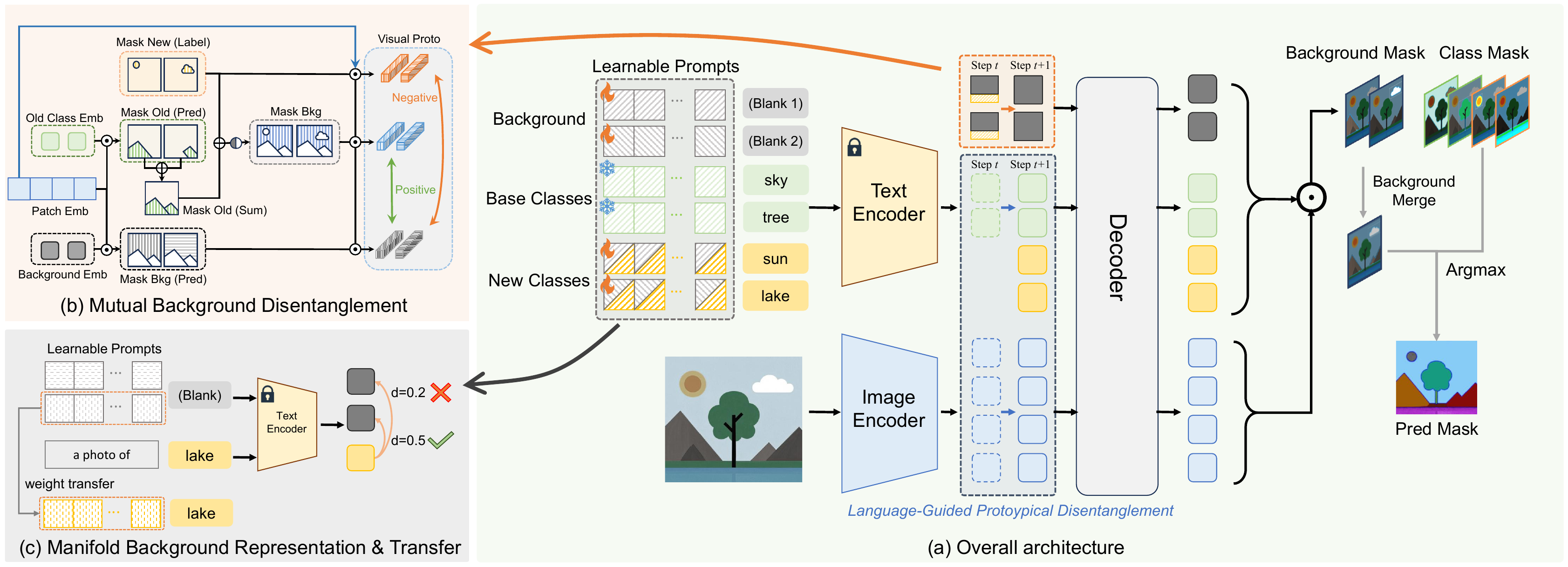}
    \caption{\textbf{Illustration of our Language-inspired Bootstrapped Disentanglement pipeline.} (a) The overall architecture of CISS, including the CLIP encoder and learnable prompts. (b) \textbf{Mutual Background Disentanglement.} CLIP-derived old class masks and ground-truth labels are used to calculate pooling-based features to achieve mutual disentanglement of the background and new classes. (c) \textbf{Manifold Background Representation with Selective Weight Transfer.} Multiple learnable prototypes are used to represent the complex semantics of the background, and the weight transfer source of the new class depends on the similarity between background and class templates. The final background mask is formed by the fusion of masks generated from multiple prototypes.}
    \label{fig:pipeline}
    \vspace{-1em}
\end{figure*}

\section{Related Work}
\label{sec:formatting}

\textbf{Class Incremental Learning. }
In class-incremental learning tasks, models are required to continually learn to recognize new classes from a sequential data stream while retaining previously learned knowledge. Data replay-based methods~\cite{DBLP:conf/cvpr/BangKY0C21,DBLP:conf/iccv/LangeT21,DBLP:journals/tnn/ZhaoWFWL22,DBLP:conf/cvpr/WangYLHL021,DBLP:conf/cvpr/ZhuZWYL21} achieve this by storing data from previous tasks or generating images of previously learned classes, allowing the model to revisit past data distributions. Network expansion-based methods~\cite{DBLP:conf/iclr/YoonYLH18,DBLP:conf/nips/XuZ18,DBLP:conf/icml/LiZWSX19,DBLP:conf/cvpr/YanX021,DBLP:conf/eccv/WangZYZ22,DBLP:conf/iclr/0001WYZ23} dynamically adjust the model's architecture or capacity during training to enhance its ability to learn new knowledge. Parameter regularization-based methods~\cite{DBLP:conf/kdd/YangZZX019,DBLP:journals/tkde/YangZZXJY23,DBLP:conf/cvpr/LeeHJK20} focus on how the model parameters should dynamically adapt when the network structure remains fixed.

\noindent\textbf{Class Incremental Semantic Segmentation. }
ILT~\cite{ILT} first introduced the CISS task.
Subsequent works~\cite{DBLP:conf/cvpr/CermelliFTCC22,DBLP:conf/eccv/LiuRZLN24,DBLP:conf/collas/RoyVCL23} investigated CISS under weak supervision.
MiB~\cite{MiB} addressed the core issue of background shift using a novel classifier initialization and a distillation loss. PLOP~\cite{PLOP} employed multi-scale pooling distillation and pseudo-labeling to retain old knowledge. SDR~\cite{SDR} reduced forgetting by shaping the latent space to maintain feature consistency and sparsity. RCIL~\cite{RCIL} utilized structural reparameterization to decouple representations of new and old knowledge.
CoMFormer~\cite{CoMFormer} introduced the continual panoramic segmentation task, proposing an adaptive distillation loss and a mask-based pseudo-label technique.
Incrementer~\cite{INC}, building on ViT architectures, added new class tokens to the decoder for incremental learning. ECLIPSE~\cite{ECLIPSE} applied visual prompt tuning to Mask2Former~\cite{Mask2Former}, significantly reducing trainable parameters.
MBS~\cite{MBS} mitigated background shift through a selective pseudo-labeling strategy and adaptive feature distillation.
Unlike prior visual-only methods, our approach leverages CLIP's multimodal information to address key challenges in continual segmentation.

\noindent\textbf{CLIP-based Semantic Segmentation. }
DenseCLIP~\cite{DenseCLIP} proposed context-aware prompting and converted CLIP's original image-text matching problem into pixel-text matching for dense prediction. MaskCLIP~\cite{MaskCLIP} directly modified CLIP's image encoder, producing reasonable segmentation results without fine-tuning. ZegCLIP~\cite{ZegCLIP} extended CLIP's zero-shot prediction ability from image-level to pixel-level.
WeCLIP~\cite{WeCLIP} explored weakly-supervised semantic segmentation, freezing CLIP's feature extractor and retaining only the trainable segmentation decoder.
ClearCLIP~\cite{ClearCLIP} highlighted the negative impact of residual connections on segmentation performance and improved segmentation by modifying the last layer of the feature extractor.
MTA-CLIP~\cite{MTACLIP} emphasized the challenge of aligning global scene representations in CLIP text embeddings with local pixel-level features, introducing a framework for mask-level visual-language alignment.
FMWISS~\cite{FMWISS} utilized the score maps from CLIP as additional signals to optimize noisy labels in weakly-supervised learning.
kNN-CLIP~\cite{kNN-CLIP} continuously embedded the visual embeddings of new classes into a database to enhance model performance in incremental open-vocabulary segmentation tasks.
Although CLIP-based segmentation models are plentiful, few have been applied to CISS tasks. Even when used, CLIP is often treated merely as an additional source of supervisory signals. Our method further integrates CLIP's generalized multimodal topological knowledge structure into the continual learning.

\section{Method}
\label{sec:method}

\subsection{Problem Definition}
\label{sec:Problem Definition}
CISS aims to simulate real-world scenarios where a model continuously learns to recognize new classes as independent tasks arrive. Typically, the training process consists of multiple timesteps, denoted as $t = 1, 2, \dots, T$. For timestep $t$, the training set can be represented as $D^t = \{(x_i^t \in \mathbb{R}^{H \times W \times 3}, y_i^t \in \mathbb{R}^{H \times W})\}_{i=1}^{N_t}$, where $N_t$ denotes the number of training images at timestep $t$, $x_i^t$ and $y_i^t$ correspond to the $i$-th image and its label map, respectively. It is important to emphasize that the classes in the label set $C^t$ (also referred to as novel classes) for timestep $t$ are disjoint from the classes in all previous timesteps $C^{1:t-1}$ (also known as old classes), \ie, $C^{1:t-1} \cap C^t = \varnothing$. If a class from $C^t$ appears in the training data of a future timestep $t'$, the corresponding regions in $D^{t'}$ are labeled as background $c_0$. After completing training at timestep $t$, the model is evaluated on test data that includes all previously seen classes, \ie, $C_{test}^t = C^1 \cup \cdots \cup C^t$. This requires the model to learn new classes effectively under the supervision of only the new classes while retaining the knowledge of previously learned classes.

\subsection{From Sparse to Dense: CLIP-based CISS}
\label{sec:CLIP for CISS}

CLIP consists of an image encoder $\mathcal{E}_v$ and a text encoder $\mathcal{E}_t$. As shown in \cref{fig:pipeline}(a), the most basic pipeline to apply CLIP to segmentation tasks~\cite{DenseCLIP} involves passing an image and the corresponding class text (\eg,``A photo of [CLS]'', where the background label can be an empty word) through their respective encoders. By removing the attention pooling layer at the end of the original image encoder~\cite{MaskCLIP}, patch features for the image are obtained. The score map for each class is derived by computing the cosine similarity between the class embeddings and the visual embeddings.

However, the original design of CLIP is tailored for image-level classification, limiting its ability to capture local details in pixel-level dense predictions~\cite{CLIPSurgery}. We follow the approach proposed by ClearCLIP~\cite{ClearCLIP}, removing the residual connections and the FFN in the final layer of the ViT, which effectively reduces segmentation noise. Regarding the generation of class embeddings, instruction tuning~\cite{VPT,L2P,POP} has been shown to be highly effective. Specifically, the input to the text encoder for the $i$-th class $\mathbf{t}_{i}^{pre}$ is
\begin{equation}
    \mathbf{t}_{i}^{pre}=[\boldsymbol{p}_i, [CLS_i]],
\end{equation}
where $\boldsymbol{p}_i \in \mathbb{R}^{N_p \times C}$ is a learnable context of length $N_p$, and $CLS_i \in \mathbb{R}^{N_{CLS} \times C}$ represents the tokenized embedding of the class name. The class embedding for each class is then obtained by passing $\mathbf{t}_i^{pre}$ through the text encoder $\mathbf{t}_i=\mathcal{E}_t(\mathbf{t}_i^{pre})$

Inspired by works like~\cite{INC}, to better facilitate the fusion of cross-modal features, we combine the class embeddings of $N+1$ classes (including the background) $\mathbf{t} = \{\mathbf{t}_{bkg}, \mathbf{t}_1, \dots, \mathbf{t}_N\}, \mathbf{t}_i \in \mathbb{R}^C$ with the visual embeddings of $M = H' \times W'$ patches outputted by the visual encoder $\mathbf{v} = \{\mathbf{v}_1, \dots, \mathbf{v}_M\}, \mathbf{v}_i \in \mathbb{R}^C$. The concatenated sequence $\{\mathbf{t}, \mathbf{v}\}$ is then passed through a transformer decoder $\mathcal{D}$ to generate the refined embeddings $\{\mathbf{t}', \mathbf{v}'\} = \mathcal{D}(\{\mathbf{t}, \mathbf{v}\})$, which are used to compute the segmentation mask for the $i$-th class:
\begin{equation}
    \boldsymbol{S}_i=\mathbf{t}_i'\mathbf{v}'.
\end{equation}

By applying operations like reshaping and upsampling, we obtain the final mask $\boldsymbol{S} \in \mathbb{R}^{(N+1) \times H \times W}$. Benefiting from the scalability of the baseline architecture, when a new class is learned, the corresponding text input is added to obtain the new class embedding. To mitigate catastrophic forgetting, we adopt the mainstream approach of using additional supervision based on pseudo-labeling~\cite{PLOP}. Specifically, we first generate masks $\boldsymbol{S}^{prev}$ for the current image using the model from the previous phase, and from these, compute the labels for old classes $\hat{Y}$. The new label set $Y'$ is then formed by combining $\hat{Y}$ with the new class labels $Y$, which is used to compute the cross-entropy loss.

\subsection{Language-Guided Protoypical Disentangle}
\label{sec:Topology-Preserving Distillation Loss}

\begin{figure}[t]
    \centering
    \includegraphics[width=\linewidth]{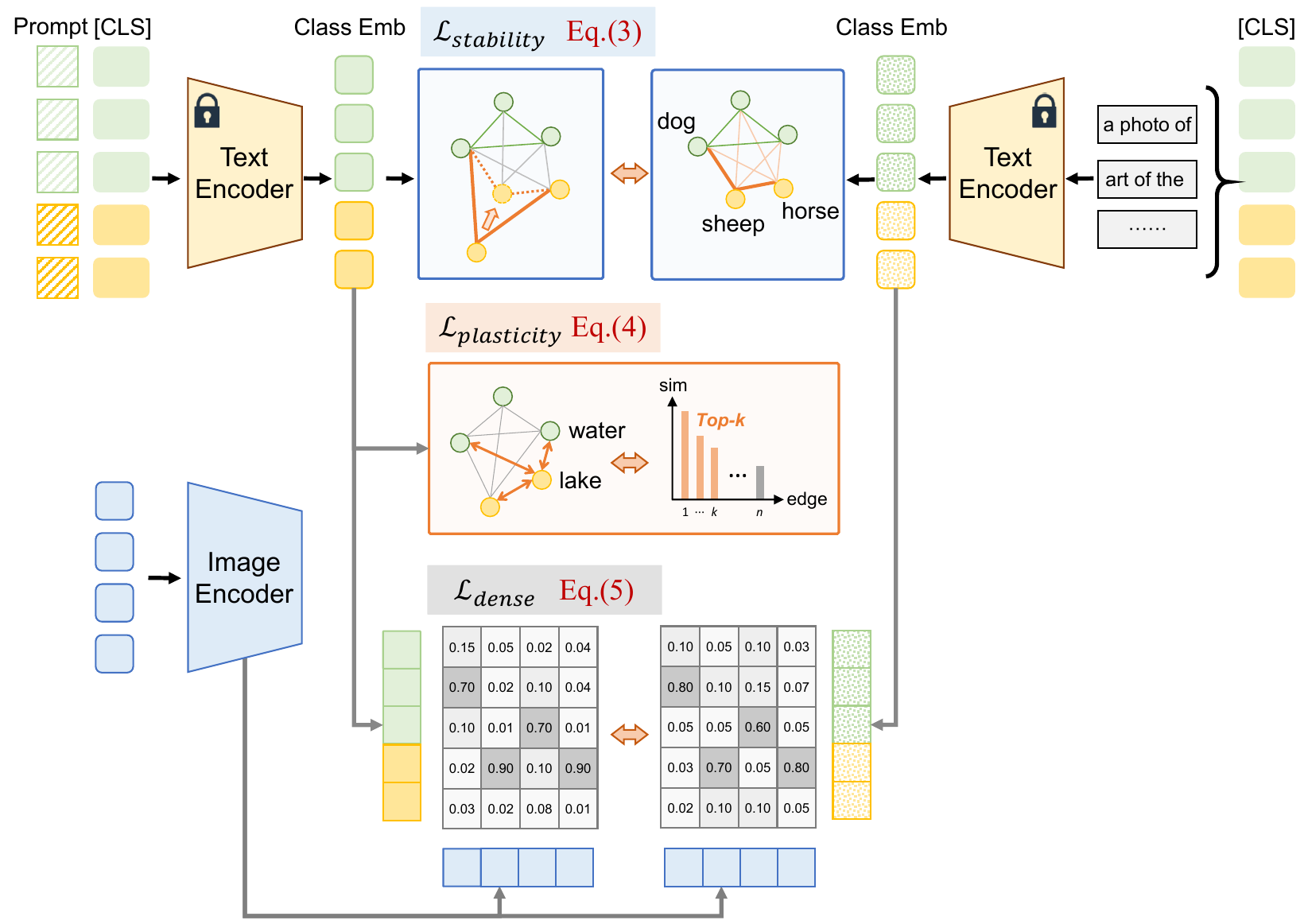}
    \caption{\textbf{Illustration of Language-guided protoypical disentangle.} Macroscopically, the topological structure of prototypes must be maintained. Microscopically, the local semantic plasticity must be ensured. The former is achieved through relationship distillation between class embeddings and templates, while the latter relies on maximum similarity constraints. Cross-modal dense learning further maintains the generalization of feature alignment.}
    \label{fig:tpd}
    \vspace{-1em}
\end{figure}

Context learning and continuous training of the visual encoder in the CLIP-based model enhance plasticity but cause the Prototype-Feature Entanglement dilemma. More learnable parameters expand the feature space, treating each incremental stage as a separate segmentation task. The prompts focus on distinguishable features within the current dataset, neglecting semantic associations between stages, which entangle prototypes and visual features. A similar issue occurs in pixel-level image-text similarity, where the absence of old class data leads to visual feature mismatches, disrupting the topological structural knowledge from CLIP.

To address this, knowledge distillation methods are widely adopted to retain the previously learned knowledge. However, existing distillation methods~\cite{INC,MBS} use the model from earlier stages as a teacher to constrain the current training, mainly preserving the consistency of feature maps and class embeddings. This approach can only transfer the output of the old model to a limited extent, without effectively decoupling it. If the structural knowledge of CLIP is treated as a graph (\cref{fig:tpd}), where class prototypes are nodes and inter-class similarity represents edges, then if the position of newly acquired nodes is poorly generalized, even if the current structure can be preserved later, the increase in class numbers will lead to misclassifications.

Therefore, the key lies in \textit{macroscopically} preserving the topological structure, and \textit{microscopically} maintaining local semantic plasticity. For the continuously updated $\mathbf{t}_i$, we use a static generalized embedding consisting of a series of manually constructed descriptions (See Appendix for details) containing explicit class names (excluding background) as \textit{templates} $\mathbf{t}^* = \{\mathbf{t}^*_1, \cdots, \mathbf{t}^*_N\}$ to be aligned. Based on this, we employ stability constraints~\cite{RKD} of class embeddings to maintain the knowledge structure learned by CLIP:
\begin{equation}
\begin{aligned}
    \mathcal{L}_{stability}&=\sum_{i,j} l_{\delta}(\psi_D(\mathbf{t}_i,\mathbf{t}_j),\psi_D(\mathbf{t}^*_i,\mathbf{t}^*_j)) \\
    &+\sum_{i,j,k} l_{\delta}(\psi_A(\mathbf{t}_i,\mathbf{t}_j,\mathbf{t}_k),\psi_A(\mathbf{t}^*_i,\mathbf{t}^*_j,\mathbf{t}^*_k)),
\end{aligned}
\end{equation}
where $l_{\delta}$ represents the Huber Loss, $\psi_D$ represents the Euclidean distance, and $\psi_A(\mathbf{t}_i, \mathbf{t}_j, \mathbf{t}_k)$ denotes the $\cos \angle \mathbf{t}_i \mathbf{t}_j \mathbf{t}_k$.

While inter-class relationships are preserved, not all aspects are beneficial at the local level. The naming conventions of datasets hinder their effectiveness (\eg,``stairs'' and ``stairway'' in the ADE20k dataset). These classes, which may appear similar in text, actually have different visual meanings. If we enforce orthogonality among all $\mathbf{t}_i$, the macroscopic topology will still be disrupted. Therefore, we propose a orthogonality constraint based on ranking scores to ensure local-plasticity, which only calculates the most similar $k$ pairs of embeddings, corresponding to the shortest $k$ edges $E=\text{Top-k}(\{ \cos(\mathbf{t}_i, \mathbf{t}_j) \})$ in the graph:
\begin{equation}
    \mathcal{L}_{plasticity} = \sum_{(i,j) \in E} (1 - \cos(\mathbf{t}_i, \mathbf{t}_j)) \cdot \mathbbm{1}_{\{ i \in C^t \}},
\end{equation}
where $\mathbbm{1}_{{i \in C^t}}$ indicates that the starting point of the edge must be a class from the current training step.

The two losses described above primarily target the language modality. To prevent the visual encoder from overfitting to the current training set, we similarly employ cross-modal dense learning based on temperature distillation, which constrains the similarity of logits between each visual patch and the class embeddings, ensuring alignment with the template embeddings:
\begin{equation}
    \mathcal{L}_{dense}=\mathcal{D}_{KL}(\text{softmax}(\boldsymbol{S}/T) \Vert \text{softmax}(\boldsymbol{S}^*/T))\cdot T^2,
\end{equation}
where $\boldsymbol{S}, \boldsymbol{S}^* \in \mathbb{R}^{N \times M}$ represent the score maps obtained by multiplying the text and visual features, and $T$ is the temperature coefficient. The complete language-guided prototypical disentanglement loss is the weighted combination of these three losses:
\begin{equation}
\mathcal{L}_{lpd}=\mathcal{L}_{stability}+\alpha\mathcal{L}_{plasticity}+\beta\mathcal{L}_{dense}.
\label{eq:lpd}
\end{equation}

\subsection{Manifold Mutual Background Disentangle}
\label{sec:Dynamic Multiple Background Representation}

The background semantics gradually change over time, leading to Background-Increment Entanglement. Mainstream methods~\cite{INC,MBS} typically rely on a single prototype to model the background, which results in the difficulty of effectively capturing and adapting to new background when a shift occurs. In fact, the background can be viewed as a collection of multiple unseen classes. Therefore, we propose the dynamic manifold background representation.

We initialize multiple mutually orthogonal learnable prompts for the background, denoted as $\boldsymbol{p}_{bkg} = \{\boldsymbol{p}_{bkg}^1, \cdots, \boldsymbol{p}_{bkg}^{n}\}, \boldsymbol{p}_{bkg}^i \in \mathbb{R}^{N_p \times C}$, and obtain $n$ background embeddings $\{\mathbf{t}_{bkg}^1, \cdots, \mathbf{t}_{bkg}^n\}$ via the encoder $\mathcal{E}_t$. These embeddings are then concatenated with the remaining embeddings $\{\mathbf{t}_1, \cdots, \mathbf{t}_N, \mathbf{v}_1, \cdots, \mathbf{v}_{M}\}$ and fed into the decoder. Based on the algorithm described in \cref{sec:CLIP for CISS}, we obtain $n$ background masks: $\boldsymbol{M}_{bkg}^i \in \mathbb{R}^{H' \times W'}, 1 \leqslant i \leqslant n$. We then derive the final background mask by taking the maximum logits at each pixel:
\begin{equation}
    \boldsymbol{M}_{bkg}'(h, w) = \max_{1 \leq i \leq n} \boldsymbol{M}_{bkg}^i(h, w),
\end{equation}
where $h$ and $w$ represent the row and column coordinates of the pixel, and $\boldsymbol{M}_{bkg}^i(h, w)$ is the value of the $i$-th background mask at pixel $(h, w)$.

The dynamic manifold representation overcomes the limitations of a single prototype by introducing multiple prototypes. These prototypes represent the semantic features of different potential classes within the background, which may include newly added classes in the current step. When initializing the prompt for a new class $c$, we select the background embedding that is most similar to the new class’s template $\mathbf{t}_c^*$ from the $n$ background embeddings to transfer the background weights:
\begin{equation}
    \boldsymbol{p}_c=\boldsymbol{p}_{bkg}^k,\quad k=\max_i \cos(\mathbf{t}_{bkg}^i,\mathbf{t}_c^*),
\end{equation}
which ensures that $\boldsymbol{p}_c$ contains the semantics of the new class while minimizing the background shift towards the new class (\cref{fig:pipeline}(c)). To further strengthen the separation between the background and the new classes, we introduce mutual background disentanglement (\cref{fig:pipeline}(b)). The core idea is to use contrastive learning to generate a reference background feature from mask pooling, then disentangle the new class from the background, thereby ensuring that the representation of the background and the new target class are as distinct as possible.

Specifically, let $N_{old}$ be the number of old classes learned at the current step, and $N_{new}$ be the number of newly added classes. For an input image, we first multiply the visual patch embeddings with the class embeddings of background and old class, followed by argmax operations to obtain the corresponding masks $\boldsymbol{S}_{bkg} = \{\boldsymbol{S}_{bkg}^1, \cdots, \boldsymbol{S}_{bkg}^n\}$ and $\boldsymbol{S}_{old} = \{\boldsymbol{S}_{old}^1, \cdots, \boldsymbol{S}_{old}^{N_{old}}\}$. By downsampling the ground-truth labels, we obtain the masks for the new classes, $\boldsymbol{S}_{new} = \{\boldsymbol{S}_{new}^1, \cdots, \boldsymbol{S}_{new}^{N_{new}}\}$. The total mask for the old classes is obtained by summing all the masks in $\boldsymbol{S}_{old}$, and the reference background mask after removing the $i$-th new class $\widehat{\boldsymbol{S}_{bkg}^i}$ is obtained by taking the union of each $\boldsymbol{S}_{new}^i$ with the summed old class mask and performing a inversion operation:
\begin{equation}
    \widehat{\boldsymbol{S}_{bkg}^i}=\neg\left( \left(\sum \boldsymbol{S}_{old}\right) \bigvee \boldsymbol{S}_{new}^i \right).
\end{equation}

By multiplying any mask $\boldsymbol{S}^i \in \mathbb{R}^{H' \times W'} = \mathbb{R}^M$ with the patch embedding $\mathbf{V} = [\mathbf{v}_1, \cdots, \mathbf{v}_M] \in \mathbb{R}^{M \times C}$, we obtain the corresponding visual features for the region. Our core goal is to make the visual features of the background region, calculated using $\mathbf{t}_{bkg}$, as dissimilar as possible to those of the new class, while making them as similar as possible to the features of the region from which the new class has been excluded. Thus, each $\boldsymbol{S}_{new}^i\mathbf{V} \in \mathbb{R}^C$ can be treated as a negative sample, and each $\widehat{\boldsymbol{S}_{bkg}^i}\mathbf{V}$ as another positive sample. All $\boldsymbol{S}_{bkg}^i\mathbf{V}$ can be treated as anchor points, from which we compute the contrastive loss:
\begin{equation}
\begin{aligned}
\mathcal{L}_{bkg} = \frac{1}{n}\sum_{i=1}^{n} \frac{1}{N_{new}} &\sum_{j=1}^{N_{new}} \Big[ \cos(\boldsymbol{S}_{bkg}^i \mathbf{V}, \boldsymbol{S}_{new}^j \mathbf{V})  + \\
&\left( 1 - \cos(\boldsymbol{S}_{bkg}^i \mathbf{V}, \widehat{\boldsymbol{S}_{bkg}^j} \mathbf{V}) \right) \Big].
\end{aligned}
\end{equation}

Through manifold representation and mutual disentanglement, the model can better handle background shift in complex environments.

\section{Experiment}

\begin{table*}[t]
\setlength{\abovecaptionskip}{0.2cm}
\belowrulesep=0pt
\aboverulesep=0pt
  \centering
 
    \scalebox{0.75}{
    \begin{tabular}{l|cccc|cccc|cccc|cccc}
    \toprule
    \multicolumn{1}{c|}{\multirow{2}[2]{*}{Method}} & \multicolumn{4}{c|}{100-50 (2 steps)} & \multicolumn{4}{c|}{50-50 (3 steps)} & \multicolumn{4}{c|}{100-10 (6 steps)} & \multicolumn{4}{c}{100-5 (11 steps)} \\
          & 1-100 & 101-150 & All   & Har.  & 1-50  & 51-150 & All   & Har.  & 1-100 & 101-150 & All   & Har.  & 1-100 & 101-150 & All   & Har. \\
    \midrule
    \multicolumn{17}{c}{\textbf{CNN-based Methods}} \\
    \midrule
    MiB~\cite{MiB}   & 40.5  & 17.2  & 32.8  & 24.1  & 45.5  & 21.0  & 29.3  & 28.7  & 38.2  & 11.1  & 29.2  & 17.2  & 36.0  & 5.7   & 26.0  & 9.8  \\
    SDR~\cite{SDR}   & 37.4  & 24.8  & 33.2  & 29.8  & 40.9  & 23.8  & 29.5  & 30.1  & 28.9  & 7.4   & 21.7  & 11.8  & -     & -     & -     & - \\
    PLOP~\cite{PLOP}  & 41.7  & 15.4  & 33.0  & 22.5  & 47.8  & 21.6  & 30.4  & 29.8  & 39.4  & 13.6  & 30.9  & 20.2  & 39.1  & 7.8   & 28.8  & 13.0  \\
    REMIND~\cite{REMIND} & 41.6  & 19.2  & 34.1  & 26.3  & 47.1  & 20.4  & 29.4  & 28.5  & 39.0  & 21.3  & 33.1  & 27.6  & -     & -     & -     & - \\
    RCIL~\cite{RCIL}  & 42.3  & 18.8  & 34.5  & 26.0  & 48.3  & 25.0  & 32.5  & 32.9  & 39.3  & 17.6  & 32.0  & 24.3  & 38.5  & 11.5  & 29.6  & 17.7  \\
    SPPA~\cite{SPPA}  & 42.9  & 19.9  & 35.2  & 27.2  & 49.8  & 23.9  & 32.5  & 32.3  & 41.0  & 12.5  & 31.5  & 19.2  & -     & -     & -     & - \\
    RBC~\cite{RBC}   & 42.9  & 21.5  & 35.8  & 28.6  & 49.6  & 26.3  & 34.2  & 34.4  & 39.0  & 21.7  & 33.3  & 27.9  & -     & -     & -     & - \\
    \midrule
    Joint (\textit{upper bound})  & 43.9  & 27.2  & 38.3  & 33.6  & 50.9  & 32.1  & 38.3  & 39.4  & 43.9  & 27.2  & 38.3  & 33.6  & 43.9  & 27.2  & 38.3  & 33.6  \\
    \midrule
    \multicolumn{17}{c}{\textbf{Transformer-based Methods}} \\
    \midrule
    MiB*~\cite{MiB}   & 46.6  & 35.0  & 42.6  & 40.0  & 52.2  & 35.6  & 41.1  & 42.3  & 43.0  & 30.8  & 38.9  & 35.9  & 40.2  & 26.6  & 35.7  & 32.0  \\
    INC*~\cite{INC}   & \underline{49.4}  & 35.6  & 44.8  & 41.4  & \textbf{56.2} & 37.8  & 43.9  & 45.2  & \underline{48.5}  & \underline{34.6}  & \underline{43.9}  & \underline{40.4}  & \textbf{46.9} & \underline{31.3}  & \underline{41.7}  & \underline{37.5}  \\
    MBS†~\cite{MBS}   & 49.3  & \underline{37.5}  & \underline{45.3}  & \underline{42.6}  & \textbf{56.2} & \underline{39.7}  & \underline{45.4}  & \underline{46.5}  & 48.1  & 34.0  & 43.7  & 39.8  & \underline{45.9}  & 30.0  & 40.6  & 36.3  \\
    \rowcolor{gray!20} Ours  & \textbf{51.3} & \textbf{38.7} & \textbf{47.1} & \textbf{44.1} & \textbf{56.2} & \textbf{40.6} & \textbf{45.8} & \textbf{47.1} & \textbf{48.7} & \textbf{34.9} & \textbf{44.1} & \textbf{40.7} & \textbf{46.9} & \textbf{31.9} & \textbf{41.8} & \textbf{38.0} \\
    \midrule
    Joint (\textit{upper bound}) & 52.9  & 42.6  & 49.5  & 47.2  & 58.9  & 44.7  & 49.5  & 50.8  & 52.9  & 42.6  & 49.5  & 47.2  & 52.9  & 42.6  & 49.5  & 47.2  \\
    \bottomrule
    \end{tabular}%
    }
     \caption{Performance comparison on ADE20k across various scenarios in \textit{overlapped} setting. CNN
and Transformer indicates the type of the backbone. * denotes results from \cite{INC}, † indicates the results reproduced using the same version of ViT as the other methods. Har. denotes the harmonic mean of the MIoU between the initial class set $C^1$ and the incremented sets $C^{2:T}$.}
  \label{tab:ade-main}
\end{table*}

\subsection{Experimental Details}

\textbf{Datasets. }
In line with the setup in~\cite{INC,MBS}, we evaluate our method using two widely recognized datasets: Pascal VOC~\cite{PascalVOC} and ADE20k~\cite{ADE}. The Pascal VOC dataset comprises 10,582 annotated training images and 1,449 testing images, spanning over 20 object classes. ADE20k consists of 20,210 images for training and 2,000 images for testing, distributed across 150 distinct classes.

\noindent\textbf{Experimental Protocols. }
To evaluate the performance of our method, we utilize a two-fold experimental setup with distinct CISS configurations: \textit{Disjoint} and \textit{Overlapped}. In both configurations, labels are assigned solely to the new classes $C_t$ introduced at each step t. At the same time, the data $D_t$ includes samples from previously learned and current step classes. Precisely, in the \textit{Disjoint} configuration, $D_t$ consists of data from the union of the old classes $C_{1:t-1}$ and the new classes $C_t$. In contrast, the \textit{Overlapped} configuration incorporates not only the current and previous classes but also data from future class sets $C_{1:t-1} \cup C_t \cup C_{t+1:T}$, representing a more challenging and realistic scenario for continuous learning. The performance under a \textit{Joint} scenario, where all classes are trained simultaneously, is also used as a best-case baseline.

To assess the incremental learning capacity, we follow a class partition strategy similar to prior works, which organizes classes based on the number of steps in the continual learning process. For example, the benchmark labeled as 15-1 (6 steps) refers to a scenario where the model is initially trained on 15 classes, then adding one new class at each subsequent step.
To ensure comparative fairness, we replace the backbone with ViT-B/16-224 (instead of 384) in the code provided by MBS~\cite{MBS} and reproduce it.

For evaluation metrics, we follow previous work by providing the average MIoU for both the basic and incremental stages, as well as the average MIoU for all classes.
We additionally provide the harmonic mean of the stage-basic and stage-new MIoU as a supplement to better reflect the trade-off between the learning performances of different stages.

\noindent\textbf{Implementation Details. }
Our method is built upon the transformer-based CLIP~\cite{CLIP}, which includes the openCLIP~\cite{OpenCLIP} pre-trained ViT-B/16~\cite{ViT} visual encoder. For the decoder, we employ a simple module consisting of two transformer decode layers, consistent with the approach used in~\cite{INC,MBS}. The input image is resized to $512 \times512$.
We use the AdamW~\cite{AdamW} optimizer with an initial learning rate of 3e-6 and a batch size of 8 for both datasets. Each training step runs for 64 epochs. For incremental sessions, the learning rate is set to 0.5 times the base rate for ADE20k and 0.1 times for Pascal VOC. To balance learning across modules, the CLIP encoder's learning rate is further reduced to 30\% of the current rate.
The learnable prompts' length is set to $N_p = 8$, with the number of background prompts $n = 4$. Before the $t$-th step in incremental learning, we freeze all prompts $C_{1:t-1}$ and perform the background weight transfer.
For \cref{eq:lpd}, we set $\alpha=1$, $\beta=0.2$.
See the Appendix for further details.

\subsection{Comparisons with the State-of-the-Arts}

\textbf{ADE20k. }
Experimental results for the overlapped setting on the ADE20k dataset are shown in \cref{tab:ade-main}. For short-step settings, our method exceeds the previous SOTA model by 1.2 (100-50) and 0.9 (50-50) on new classes, demonstrating the stronger baseline brought by the inclusion of additional textual information. In the two other long-step settings, our method shows an even greater margin. One major characteristic of the ADE20k dataset is its large number of classes, complex inter-class relationships, and significant background shift, making it prone to confusion during continual learning. Thanks to bootstrapped disentanglement on class embeddings, our method performs consistently across all settings, effectively controlling the phenomenon of forgetting.

\begin{table}[t]
\setlength{\abovecaptionskip}{0.2cm}
\belowrulesep=0pt
\aboverulesep=0pt
  \centering
  
    \scalebox{0.9}{
    \begin{tabular}{c|ccc}
    \toprule
    Method & 1-15  & 16-20 & All \\
    \midrule
    INC   & 79.6  & 59.6  & 75.6  \\
    INC+CLIP & 80.7 \textcolor{teal}{(+1.1)}  & 58.6 \textcolor{orange}{(-1.0)}  & 76.1 \textcolor{teal}{(+0.5)}  \\
    \midrule
    MBS   & 80.9  & 64.9  & 77.6  \\
    MBS+CLIP & 81.0 \textcolor{teal}{(+0.1)}  & 64.3 \textcolor{orange}{(-0.6)}  & 76.8 \textcolor{orange}{(-0.8)}  \\
    \midrule
    Ours (Fix)  & 78.9  & 49.0 & 72.1  \\
    \midrule
    Ours  & \textbf{81.9}  & \textbf{66.6}  & \textbf{78.1}  \\
    \bottomrule
    \end{tabular}
    }
    \caption{Comparison of the methods combined with CLIP in the Pascal VOC 15-1 \textit{overlapped} setting.}
  \label{tab:clip}
\vspace{-15pt}
\end{table}

\noindent\textbf{Pascal VOC.}
Comprehensive experimental results on Pascal VOC are presented in \cref{tab:voc-main}.
While improving accuracy for new classes, we exhibit less forgetting on old classes. Notably, in multi-step scenarios (15-1), where the data distribution and background semantics continually shift, our proposed method surpasses the previous SOTA by 1.4 (disjoint) and 1.7 (overlapped) on new classes. When compared with the joint results, we are closer to the theoretical upper bound. 
It further emphasizes the importance of retaining original CLIP topological knowledge and dynamically constraining the background to balance model stability and plasticity.

\noindent\textbf{Is All the Credit Owed to CLIP?}
Since we use the CLIP backbone, which leverages more training data than the ImageNet-pretrained ViT used by other methods, we conducted supplementary experiments to eliminate the potential interference of this factor on the experimental results. We replaced the backbones of MBS~\cite{MBS} and INC~\cite{INC} with CLIP, keeping the class embedding initialization consistent with ours (considering the background as a regular class). All other settings remained the same. From the results in \cref{tab:clip}, it is evident that the inclusion of CLIP does slightly improve the baseline performance (indicated by \textcolor{teal}{green} values), but the forgetting phenomenon still exists and is even more severe than in the original method (indicated by \textcolor{orange}{orange} values). The cause of this phenomenon lies in our retention of the CLIP visual encoder's training. Without the use of additional regularization methods, its visual and language features misalign as learning progresses, reducing accuracy. Furthermore, if the issue is attempted to be circumvented by freezing the visual encoder, the model's learning performance significantly deteriorates (second-to-last row).

\begin{table*}[t]
\setlength{\abovecaptionskip}{0.2cm}
\setlength{\tabcolsep}{2.44pt} 
\belowrulesep=0pt
\aboverulesep=0pt
  \centering
  
    \scalebox{0.75}{

    \    \begin{tabular}{l|cccc|cccc|cccc|cccc|cccc|cccc}
    \toprule
    \multicolumn{1}{c|}{\multirow{3}[4]{*}{Method}} & \multicolumn{8}{c|}{19-1 (2 steps)}                           & \multicolumn{8}{c}{15-5   (2 steps)}                          & \multicolumn{8}{c}{15-1 (6 steps)} \\
\cmidrule{2-25}          & \multicolumn{4}{c|}{Disjoint} & \multicolumn{4}{c|}{Overlapped} & \multicolumn{4}{c|}{Disjoint} & \multicolumn{4}{c|}{Overlapped} & \multicolumn{4}{c|}{Disjoint} & \multicolumn{4}{c}{Overlapped} \\
          & 1-19  & 20    & All   & Har.  & 1-19  & 20    & All   & Har.  & 1-15  & 16-20 & All   & Har.  & 1-15  & 16-20 & All   & Har.  & 1-15  & 16-20 & All   & Har.  & 1-15  & 16-20 & All   & Har. \\
    \midrule
    \multicolumn{25}{c}{\textbf{CNN-based Methods}} \\
    \midrule
    EWC~\cite{EWC}   & 23.2  & 16.0  & 22.9  & 23.2  & 26.9  & 14.0  & 26.3  & 18.4  & 26.7  & 37.7  & 29.4  & 31.3  & 24.3  & 35.5  & 27.1  & 28.9  & 0.3   & 4.3   & 1.3   & 0.6   & 0.3   & 4.3   & 1.3   & 0.6  \\
    ILT~\cite{ILT}   & 69.1  & 16.4  & 66.4  & 26.5  & 67.8  & 10.9  & 65.1  & 18.8  & 63.2  & 39.5  & 57.3  & 48.6  & 67.1  & 39.2  & 60.5  & 49.5  & 3.7   & 5.7   & 4.2   & 4.5   & 8.8   & 8.0   & 8.6   & 8.4  \\
    MiB~\cite{MiB}   & 69.6  & 25.6  & 67.4  & 37.4  & 71.4  & 23.6  & 69.2  & 35.5  & 71.8  & 43.3  & 64.7  & 54.0  & 76.4  & 50.0  & 70.1  & 60.4  & 46.2  & 12.9  & 37.9  & 20.2  & 34.2  & 13.5  & 29.3  & 19.4  \\
    SDR~\cite{SDR}   & 69.9  & 37.3  & 68.4  & 48.6  & 69.1  & 32.6  & 67.4  & 44.3  & 73.5  & 47.3  & 67.2  & 57.6  & 75.4  & 52.6  & 69.9  & 62.0  & 59.2  & 12.9  & 48.1  & 21.2  & 44.7  & 21.8  & 39.2  & 29.3  \\
    PLOP~\cite{PLOP}  & 75.4  & 38.9  & 73.6  & 51.3  & 75.4  & 37.4  & 73.5  & 50.0  & 71.0  & 42.8  & 64.3  & 53.4  & 75.7  & 51.7  & 70.1  & 61.4  & 57.9  & 13.7  & 46.5  & 22.2  & 65.1  & 47.8  & 62.7  & 55.1  \\
    RECALL~\cite{RECALL} & 65.2  & 50.1  & 65.8  & 56.7  & 67.9  & 53.5  & 68.4  & 59.8  & 66.3  & 49.8  & 63.5  & 56.9  & 66.6  & 50.9  & 64.0  & 57.7  & 66.6  & 44.9  & 62.1  & 53.6  & 65.7  & 47.8  & 62.7  & 55.3  \\
    REMIND~\cite{REMIND} & -     & -     & -     & -     & 76.5  & 32.3  & 74.4  & 45.4  & -     & -     & -     & -     & 76.1  & 50.7  & 70.1  & 60.9  & -     & -     & -     & -     & 68.3  & 27.2  & 58.5  & 38.9  \\
    RCIL~\cite{RCIL}  & -     & -     & -     & -     & -     & -     & -     & -     & 75.0  & 42.8  & 67.3  & 54.5  & 78.8  & 52.0  & 72.4  & 62.7  & 66.1  & 18.2  & 54.7  & 28.5  & 70.6  & 23.7  & 59.4  & 35.5  \\
    SPPA~\cite{SPPA}  & 75.5  & 38.0  & 73.7  & 50.6  & 76.5  & 36.2  & 74.6  & 49.1  & 75.3  & 48.7  & 69.0  & 59.1  & 78.1  & 52.9  & 72.1  & 63.1  & 59.6  & 15.6  & 49.1  & 24.7  & 66.2  & 23.3  & 56.0  & 34.5  \\
    RBC~\cite{RBC}   & 76.4  & 45.8  & 75.0  & 57.3  & 77.3  & 55.6  & 76.2  & 64.7  & 75.1  & 49.7  & 69.9  & 59.8  & 76.6  & 52.8  & 70.9  & 62.5  & 61.7  & 19.5  & 51.6  & 29.6  & 69.5  & 38.4  & 62.1  & 49.5  \\
    \midrule
    Joint (\textit{upper bound}) & 77.4  & 78.0  & 77.4  & 77.7  & 77.4  & 78.0  & 77.4  & 77.7  & 79.1  & 72.6  & 77.4  & 75.7  & 79.1  & 72.6  & 77.4  & 75.7  & 79.1  & 72.6  & 77.4  & 75.7  & 79.1  & 72.6  & 77.4  & 75.7  \\
    \midrule
    \multicolumn{25}{c}{\textbf{Transformer-based Methods}} \\
    \midrule
    MiB*~\cite{MiB}   & 80.6  & 45.2  & 79.6  & 57.9  & 79.9  & 47.7  & 79.1  & 59.7  & 75.0  & 59.9  & 72.3  & 66.6  & 78.6  & 63.1  & 75.6  & 70.0  & 66.7  & 26.3  & 58.3  & 37.7  & 72.6  & 23.1  & 61.7  & 35.0  \\
    RBC*~\cite{RBC}   & 80.9  & 42.1  & 79.7  & 55.4  & 80.2  & 38.8  & 79.0  & 52.3  & 77.7  & 59.1  & 74.0  & 67.1  & 78.9  & 62.0  & 75.5  & 69.4  & 69.0  & 28.4  & 60.5  & 40.2  & 75.9  & 40.2  & 68.2  & 52.6  \\
    INC*~\cite{INC}   & \textbf{82.4} & 64.2  & \textbf{82.2} & 72.2  & \textbf{82.5} & 61.0  & \textbf{82.1} & 70.1  & \textbf{81.6} & 62.2  & 77.6  & 70.6  & 82.5  & 69.3  & 79.9  & 75.3  & \textbf{81.4} & 57.1  & \textbf{76.3} & 67.1  & 79.6  & 59.6  & 75.6  & 68.2  \\
    MBS†~\cite{MBS}   & 81.4  & \underline{69.3}  & \underline{81.4}  & \underline{74.9}  & 81.9  & \underline{66.1}  & \underline{81.7}  & \underline{73.2}  & 80.8  & \underline{66.9}  & \textbf{78.2} & \underline{73.2}  & \underline{83.1}  & \underline{72.4}  & \underline{80.4}  & \underline{77.4}  & 78.5  & \underline{60.9}  & \underline{74.9}  & \underline{68.6}  & \underline{80.9}  & \underline{64.9}  & \underline{77.6}  & \underline{72.0}  \\
    \rowcolor{gray!20}  Ours  & \underline{81.7}  & \textbf{70.1} & 81.1  & \textbf{75.5} & \underline{82.2}  & \textbf{70.0} & 81.6  & \textbf{75.6} & \underline{81.2}  & \textbf{67.7} & \underline{78.0}  & \textbf{73.8} & \textbf{83.2} & \textbf{73.6} & \textbf{80.8} & \textbf{78.1} & \underline{81.0}  & \textbf{62.3} & \textbf{76.3} & \textbf{70.4} & \textbf{81.9} & \textbf{66.6} & \textbf{78.1} & \textbf{73.5} \\
    \midrule
    Joint (\textit{upper bound}) & 83.0  & 83.2  & 83.0  & 83.1  & 83.0  & 83.2  & 83.0  & 83.1  & 83.6  & 81.3  & 83.0  & 82.4  & 83.6  & 81.3  & 83.0  & 82.4  & 83.6  & 81.3  & 83.0  & 82.4  & 83.6  & 81.3  & 83.0  & 82.4  \\
    \bottomrule
    \end{tabular}
    }
    \caption{Performance comparison on Pascal VOC under various scenarios. * denotes results from \cite{INC}, † indicates the results reproduced using the same version of ViT as the other methods. Har. denotes the harmonic mean of the MIoU between the initial class set $C^1$ and the incremented sets $C^{2:T}$.}
  \label{tab:voc-main}
\end{table*}

\begin{table}[t]
\setlength{\abovecaptionskip}{0.2cm}
\setlength{\belowcaptionskip}{-0.2cm}
\belowrulesep=0pt
\aboverulesep=0pt
  \centering
     \scalebox{0.9}{
    \begin{tabular}{cccc|ccc}
    \toprule
    Prompt & LPD  & Manifold & MBD   & 1-15  & 16-20 & All \\
    \midrule
          &       &       &       & 75.9  & 48.1  & 68.9 \\
    $\checkmark$     &       &       &       & 78.8  & 62.8  & 74.8 \\
    $\checkmark$     & $\checkmark$     &       &       & 81.4  & 65.5  & 77.4 \\
    $\checkmark$     & $\checkmark$     & $\checkmark$     &       & 81.6  & 66.1  & 77.7 \\
    $\checkmark$     & $\checkmark$     & $\checkmark$     & $\checkmark$     & 81.9  & 66.6  & 78.1 \\
    \bottomrule
    \end{tabular}
    }
    \caption{Ablation study for each component on Pascal VOC 15-1 overlapped setting. Prompt, Manifold denote learnable prompts, and manifold background representation, respectively.}
  \label{tab:ablation}
  \vspace{-5pt}
\end{table}

\subsection{Ablation Studies}

\textbf{Component Analysis. }
To assess the effectiveness of each module, we conduct ablation experiments on the 15-1 overlapped setting of Pascal VOC.
As shown in \cref{tab:ablation}, the first analysis focuses on the baseline that relies solely on the knowledge distillation from the output of prev model without any learnable prompts. Although the zero-shot generalization ability of CLIP is impressive, it does not adapt well to pixel-level segmentation tasks.
After adding prompt tuning, the model's plasticity is significantly improved (with a 5.9-point increase on \textit{all}). However, it still faces more severe forgetting than the single-modal ViT (2.1 points lower than MBS on 15-5), which is mainly due to the entanglement of CLIP's original topological structure caused by the continuous training of the model, as discussed in \cref{sec:Topology-Preserving Distillation Loss}. Therefore, when language-guided protoypical disentangle is further applied, the forgetting of old classes is mitigated (with a 2.6-point improvement on \textit{all}). To better represent the background, we set up multiple learnable prototypes. We use text-supervised class templates to selectively initialize the prompts for new classes, further enhancing the model's performance (with a 0.3-point increase on \textit{all}). Building on this, to promote the separation of background and new classes, we designed mutual background disentanglement, which effectively improved the accuracy of new classes (by 0.5 points).

\begin{figure}[t]
    \centering
    \includegraphics[width=\linewidth]{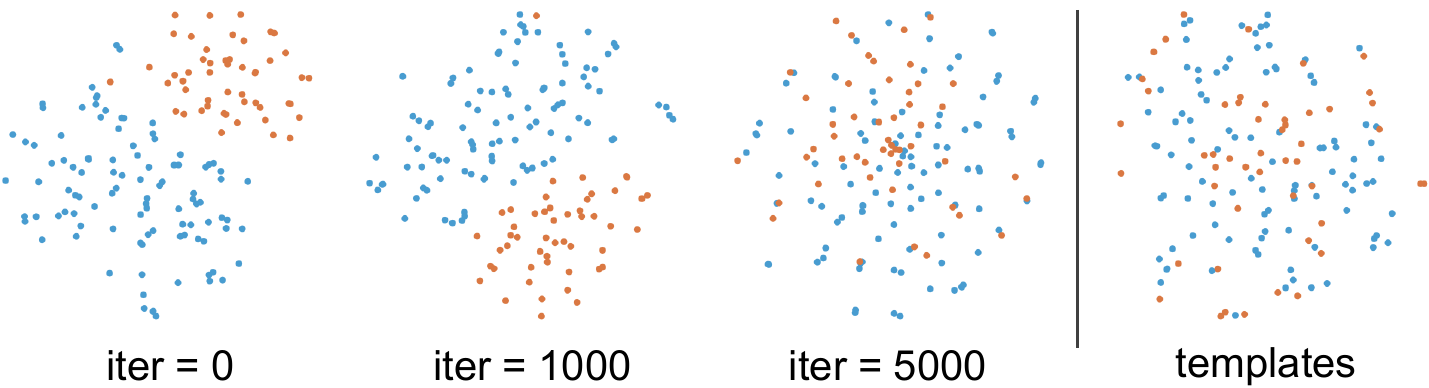}
    \caption{Visualization of class embeddings across different iters on ADE20k 100-50 setting. Under the guidance of the templates, with the increase of iterations, the prototype topology between new and basic classes gradually restores a generalized stable state.}
    \label{fig:class embeddings}
    \vspace{-1em}
\end{figure}

\noindent\textbf{Analysis of Language-guided Protoypical Disentanglement.}
\cref{fig:class embeddings} shows the t-SNE visualization results of the class template features constructed on the ADE20k 100-50 dataset, along with the CLIP class embeddings during training on new classes. The blue and red points represent the old and new classes, respectively. Due to the initialization of the prompts for the new classes with background weights, a noticeable gap in data distribution exists at the beginning between the new and old classes. As training progresses, the distillation loss drives the embeddings to align with the templates gradually, ensuring their generalizability in the feature space.

\begin{figure}[t]
    \centering
    \includegraphics[width=\linewidth]{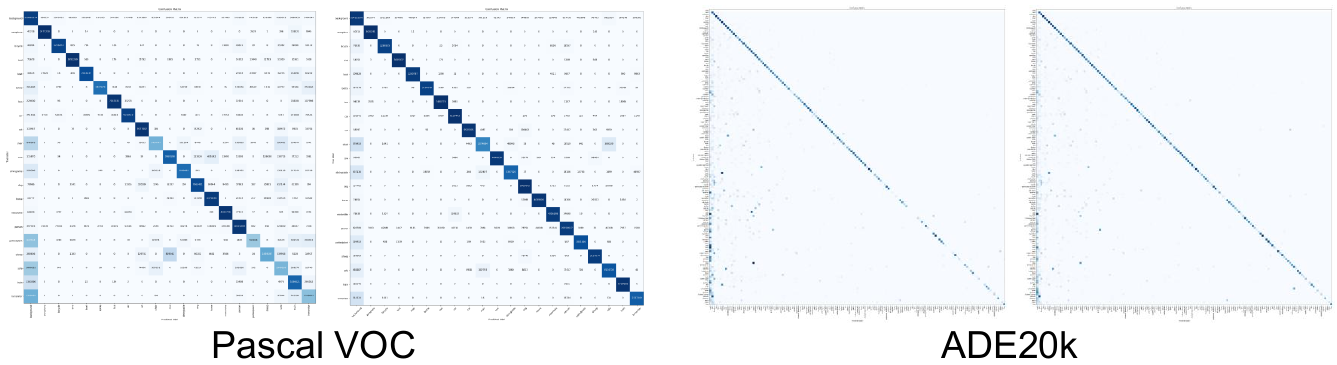}
    \caption{Visualization of the confusion matrix. After applying mutual background disentanglement, the phenomenon of background-new class overlap (left) is improved (right).
    }
    \label{fig:confusion matrices}
    \vspace{-15pt}
\end{figure}

\noindent\textbf{Analysis of Manifold Mutual Background Disentanglement.}
\cref{fig:confusion matrices} presents the visualization results of the confusion matrices for two datasets. On the left, the baseline exhibits an apparent phenomenon of new classes shifting towards the background (with the leftmost column becoming darker). After applying the series of background representation optimization methods proposed in \cref{sec:Dynamic Multiple Background Representation}, the result on the right shows significant improvement.

\section{Conclusion}

This paper abstracts the CISS method into visual feature extraction and prototype-feature matching, addressing the core issue of catastrophic semantic entanglement. We propose the language-inspired bootstrapped disentanglement framework, which guides the model to learn disentangled features using pre-trained CLIP's prior class semantics. Language-guided prototypical disentanglement uses hand-crafted textual features as class templates to disentangle new prototypes, while manifold mutual background disentanglement leverages multiple learnable prompts and mask-pooling-based contrast to disentangle backgrounds and new classes. Our method outperforms the state-of-the-art on two datasets.
\section*{Acknowledgments}

This work is partially supported by grants from the National Natural Science Foundation of China (No. 62132002 and No. 62202010), Beijing Nova Program (No.20250484786), and the Fundamental Research Funds for the Central Universities.

{
    \small
    \bibliographystyle{ieeenat_fullname}
    \bibliography{main}
}

\clearpage
\setcounter{page}{1}
\maketitlesupplementary
\renewcommand\thesection{\arabic{section}}
\setcounter{section}{0}

\section{Model Details}
\label{sec:rationale}

\subsection{Visual Encoder}

Since the original version of CLIP~\cite{CLIP,OpenCLIP} was trained on classification tasks at the image level, it cannot be directly applied to segmentation tasks. To address this, we synthesized insights from existing methods and implemented the following improvements (all encoders are based on the transformer architecture):

1. Following MaskCLIP~\cite{MaskCLIP}, we removed the average pooling in the last layer of the CLIP visual encoder ViT, which allows us to obtain dense features.

2. Following ClearCLIP~\cite{ClearCLIP}, we directly removed the feedforward neural network and residual connections from the last layer of ViT. Additionally, we replaced the attention mechanism in the final layer with v-v attention.

3. Inspired by the concept of multi-scale feature extraction~\cite{SPP}, we first extracted features from different layers of the CLIP visual encoder (specifically, the 4th, 6th, 8th, and 12th layers), concatenated them along the feature dimension, and then used convolution operations to restore the previous dimensions. This feature was then used as input to the decoder.

\subsection{Text Encoder}

To obtain class templates, we first extracted the corresponding language features from multiple text descriptions containing the class information and then computed the average of the multiple features for each class. The descriptions we used include:
 \begin{itemize}
 \item A photo of a \{\}.
 \item A snapshot of a \{\}.
 \item A bad photo of the \{\}.
 \item A clean origami \{\}.
 \item A photo of the large \{\}.
 \item A \{\} in a video game.
 \item Art of the \{\}.
 \item A photo of the small \{\}.
 \item A \{\} in the scene.
\end{itemize}

\begin{table}[t]
\setlength{\tabcolsep}{5pt}
\belowrulesep=0pt
\aboverulesep=0pt
\setlength{\belowcaptionskip}{0.0cm}
\centering
\caption{Computational and performance comparison. Our LBD method significantly outperforms DenseCLIP with only a minor increase in computational cost. Notably, key components of LBD are training-only and do not affect inference speed.}
\label{tab:computation_analysis}

\scalebox{0.9}{
 \begin{tabular}{c|cccc}
    \toprule
    Method & \thead{DenseCLIP \\ (Zero-shot)} & \thead{DenseCLIP \\ (Continual-train)}  & \thead{LBD \\ (Ours)} & Joint \\
    \midrule
    VOC 15-1 All & 61.2  & 68.7    & 78.1  & \textcolor{Gray}{83.0} \\
    Params (M) & 105.3 & 105.3  & 121.1 & \textcolor{Gray}{-} \\
    GFLOPs & 143.8 & 143.8 & 148.2  & \textcolor{Gray}{-} \\
    \bottomrule
    \end{tabular}
    }
\end{table}

\section{Analysis of Computational Cost}
\label{sec:appendix_compute}
In the domain of Continual Learning (CL), model efficiency is as crucial as performance. To provide a clear perspective on the computational overhead of our proposed Language-inspired Bootstrapped Disentanglement (LBD) method, we conduct a comparative analysis against DenseCLIP \cite{DenseCLIP}, a strong baseline that adapts the CLIP model for dense prediction tasks. This analysis is crucial for contextualizing the performance gains documented in the main paper.

Our evaluation, summarized in Table \ref{tab:computation_analysis}, focuses on three key metrics: performance (mIoU on VOC 15-1 All), model size (Parameters), and computational load (GFLOPs). We assess DenseCLIP in both its zero-shot capacity and after being continually trained on the same CISS task protocol as our LBD. The results reveal that LBD achieves a mIoU of 78.1, substantially outperforming the continually-trained DenseCLIP (68.7).
Regarding the computational budget, LBD exhibits only a marginal increase in complexity. The GFLOPs increase from 143.8 to 148.2, a modest rise of approximately 3\%. This slight overhead is primarily attributed to the learnable prompts and the lightweight adapter module. The increase in parameters from 105.3M to 121.1M similarly reflects the inclusion of these task-specific components.

Crucially, it is important to note that our core architectural innovations, such as the Language-guided Prototypical Disentanglement (LPD) module, are designed to operate \textbf{exclusively during the training phase}. These components guide the model's feature space to form a disentangled semantic structure but are detached for inference. Consequently, they introduce no additional computational burden at deployment time. Given the substantial performance improvements, especially in challenging multi-step CISS scenarios, we conclude that the minor increase in training computation is a well-justified trade-off.

\section{Exploration of PEFT}
\label{sec:appendix_peft}
The advent of large-scale pre-trained models has spurred the development of Parameter-Efficient Fine-Tuning (PEFT) methods, which aim to adapt these models to downstream tasks by updating only a small fraction of their parameters. To assess the feasibility of this paradigm for Class-Incremental Semantic Segmentation (CISS), we conducted an ablation study investigating different PEFT strategies within our LBD framework.

While our primary experiments configure the visual encoder (CLIP-ViT) as fully trainable to maximize adaptation, integrating PEFT is indeed a feasible alternative. Our study, presented in Table \ref{tab:peft_analysis}, explores the impact of selectively training different components: \ding{202} the learnable prompts introduced in Section \textcolor{red}{3.2}, \ding{203} a convolution-based adapter module placed after the encoder, and \ding{204} the full image encoder itself.

The results yield a clear insight: while PEFT approaches show promise, they currently do not match the performance of full fine-tuning for the demanding task of CISS. Training only the prompts (\ding{202}) or the adapter (\ding{203}) results in mIoU scores of 64.8 and 66.9, respectively. Combining these two PEFT techniques (\ding{202}+\ding{203}) improves the score to 72.1. However, this is still considerably lower than the 78.1 mIoU achieved when the visual encoder is fully trained (\ding{202}+\ding{203}+\ding{204}).

This performance gap suggests that adapting the vision-language model to a dense, pixel-level prediction task like semantic segmentation requires more than just peripheral modifications. The supervised signal from pixel-level annotations appears crucial for fundamentally reshaping the features within the visual backbone, an adaptation that cannot be fully achieved when the encoder is frozen. We conclude that while PEFT offers a promising avenue for reducing the training cost of CISS, future work is needed to develop more sophisticated methods that can bridge this performance gap.

\begin{table}[h]
\setlength{\tabcolsep}{5pt}
\belowrulesep=0pt
\aboverulesep=0pt
\centering
\caption{Ablation study on integrating PEFT methods within our framework on Pascal VOC 15-1 \textit{All}. We evaluate training different combinations of: \ding{202} Prompts, \ding{203} Adapter, and \ding{204} the full Image Encoder. Full fine-tuning of the encoder remains essential for achieving top performance.}
\label{tab:peft_analysis}

    \begin{tabular}{c|ccccc}
    \toprule
    Reference & \multicolumn{5}{c}{ \thead{\ding{202} Prompts (Sec.\textcolor{red}{3.2})  \ding{203} Adapter (after encoder) \\  \ding{204} Image Encoder (CLIP-ViT)}} \\
    \midrule
     Trainable & \ding{202}     & \ding{203}     & \ding{202}\ding{203}    & \ding{202}\ding{204}    & \ding{202}\ding{203}\ding{204} \\
    \midrule
    VOC 15-1 All & 64.8  & 66.9 & 72.1  & 77.4  & 78.1 \\
    \bottomrule
    \end{tabular}
\end{table}

\section{Limitations}

Our method relies on explicit class names, and when only images and numeric labels are available in the dataset, we are unable to leverage textual information. Moreover, due to the limitations of CLIP's pretraining data, CLIP fails to capture the semantic relationships between rare concepts and other classes, thus restricting the effectiveness of our method. Future work could focus on text supervision methods more suitable for incremental learning and cross-modal feature interaction.

\end{document}